# LeafTrackNet: A deep learning framework for robust leaf tracking in top-down plant phenotyping

Shanghua Liu [a,e,*], Majharulislam Babor [a], Christoph Verduyn [b], Breght Vandenberghe [b], Bruno Betoni Parodi [c], Cornelia Weltzien [a,d], Marina M.-C. Höhne [a,e,*]

[a] Leibniz Institute for Agricultural Engineering and Bioeconomy, Germany
[b] BASF Belgium Coordination Center CommV, Belgium
[c] BASF Metabolome Solutions GmbH, Germany
[d] Technical University Berlin, Germany
[e] University of Potsdam, Germany

ARTICLE INFO

Dataset link: https://github.com/shl-shawn/LeafTrackNet

ABSTRACT

High-resolution phenotyping at the level of individual leaves offers fine-grained insights into plant development and stress responses. However, the full potential of accurate leaf tracking over time remains largely unexplored due to the absence of robust tracking methods, particularly for structurally complex crops such as canola. Existing plant-specific tracking methods are typically limited to small-scale species or rely on constrained imaging conditions. In contrast, generic multi-object tracking (MOT) methods are not designed for dynamic biological scenes. Progress in the development of accurate leaf tracking models has also been hindered by a lack of large-scale datasets captured under realistic conditions. In this work, we introduce **CanolaTrack**, a new benchmark dataset comprising 5704 RGB images with 31,840 annotated leaf instances collected from 184 canola plants during their early growth stages. To enable accurate leaf tracking over time, we introduce **LeafTrackNet**, an efficient framework that combines a YOLOv10-based leaf detector with a MobileNetV3-based embedding network. During inference, leaf identities are maintained over time through an embedding-based memory association strategy. When trained directly on each target dataset without prior CanolaTrack fine-tuning, LeafTrackNet achieves HOTA scores of 88.03, 87.33, and 74.20 on CanolaTrack, KOMATSUNA, and MSU-PID, respectively, outperforming the corresponding second-best methods by 8.35, 4.94, and 1.62 HOTA points. All results are reported as mean ± standard deviation over three independent runs, using fixed inference hyperparameters across datasets without per-dataset tuning. This work provides a new benchmark for leaf-level tracking under realistic conditions and introduces CanolaTrack, which, to the best of our knowledge, is the largest leaf-tracking dataset for agricultural crops. Our code and dataset are publicly available at GitHub.

## 1. Introduction

Automated plant phenotyping is a key element in modern agriculture, offering scalable and precise tools to monitor plant growth, enabling early detection of stress responses, and accelerating crop improvement through data-driven insights. While traditional whole-plant phenotyping provides valuable information for assessing overall biomass or yield, it often does not reveal critical intra-plant dynamics, such as subtle organ-level changes that can indicate developmental or physiological shifts.

To overcome this limitation, organ-level phenotyping has emerged as a powerful alternative, enabling detailed monitoring of individual plant structures. Among these, leaves are particularly informative because they are primary sites of photosynthesis and gas exchange and often exhibit early visible symptoms of biotic and abiotic stress caused by drought, nutrient deficiency, or pathogen infection (Yan et al., 2016; Jeong et al., 2025; Zhang et al., 2024). Leaf-level traits — such as emergence timing, growth rate, morphological changes, and senescence — are not only sensitive indicators of plant health but also carry phenotypic signals for genotype evaluation and stress resilience screening (Cai et al., 2016). By temporally tracking individual leaves, we obtain continuous per-leaf time series that characterize morphological descriptors such as area, length, and curvature. These trajectories support estimation of instantaneous growth rate, acceleration, and the timing of key developmental transitions, including leaf emergence, the expansion plateau, and the onset of senescence, which are widely used as indicators of plant vigor and stress responses (Das Choudhury et al.,

* Corresponding authors at: Leibniz Institute for Agricultural Engineering and Bioeconomy, Germany.
*E-mail addresses:* sliu@atb-potsdam.de (S. Liu), mhoehne@atb-potsdam.de (M.M.-C. Höhne).

Open access funding enabled and organized by Projekt DEAL.

2018). Moreover, leaf-resolved temporal analysis can reveal treatment-driven dynamics, such as responses to nutrient availability, that may be attenuated or obscured in whole-plant summaries (Swartz et al., 2023). The ability to perform detailed and automated morphological evaluation at leaf level holds considerable promise for advancing precision agriculture and crop resilience.

In this context, *Brassica napus* (canola) is of particular interest due to its economic and ecological value. It is widely cultivated for edible oil, animal feed, and biofuel production, and plays a key role in crop rotation systems for improving soil health (Correndo et al., 2024; Lassoued et al., 2025). However, canola's complex rosette architecture, dynamic growth patterns, large variations in leaf size and shape, and frequent occlusions pose considerable challenges for automated organ-level phenotyping—especially when targeting individual leaf tracking throughout plant development. Addressing these challenges requires imaging modalities and tracking methods that can effectively capture complex leaf structures over time.

Top-down RGB imaging has become a popular approach in leaf-level plant phenotyping, particularly for rosette-stage crops like canola and Arabidopsis (*Arabidopsis thaliana*). This modality offers a compelling trade-off between cost, scalability, and resolution, without the complexity of 3D reconstruction or multi-angle capture systems (Daviet et al., 2022). Other imaging modalities such as chlorophyll fluorescence (Jurado-Ruiz et al., 2024), depth sensors (Uchiyama et al., 2017), and infrared cameras (Cruz et al., 2016) can provide additional structural or physiological information. However, these systems are often expensive and require specialized setups and controlled lighting, making them less practical for long-term or large-scale deployment. In contrast, RGB imagery offers a practical and scalable solution, encoding rich visual information, including color, texture, and structure, which are crucial for accurate leaf-level identification and temporal tracking. However, leveraging RGB images for long-term leaf tracking remains challenging. Leaves with similar shape and appearance frequently occlude one another, new leaves emerge while older ones senesce, and rotational effects due to pot movement introduce orientation inconsistencies. These dynamics over time can lead to identity switches, drift, and tracking fragmentation, resulting in poor tracking performance if not explicitly handled.

To address such challenges, several plant-specific methods have been proposed. LeTra (Jurado-Ruiz et al., 2024) performs leaf instance tracking via IoU-based mask matching, and Plant-Doctor (Montagut-Marquès et al., 2025) uses a lightweight CNN to learn appearance embeddings for associating diseased leaf detections. While promising, these methods often struggle with long-term identity preservation and occlusions, especially in complex and dynamic crops such as canola. These challenges indicate that reliable leaf tracking requires a representation that is less dependent on instantaneous geometric correspondence and better able to preserve leaf identity across growth-induced appearance changes. In particular, geometric cues such as centroid position, leaf size, and mask overlap can become unstable when leaves expand, deform, overlap, disappear, or shift under pot rotation. Conversely, state-of-the-art multi-object tracking methods for general computer vision, including ByteTrack (Zhang et al., 2022), BoT-SORT (Aharon et al., 2022), MOTRv2 (Zhang et al., 2023), and MOTIP (Gao et al., 2025), have demonstrated strong performance in real-world applications such as pedestrian or vehicle tracking. However, these models do not transfer directly to the plant domain because they rely on assumptions such as rigid-body motion, stable geometry, and consistent visual features, which are frequently violated in plant settings due to nonlinear leaf growth, self-occlusion, and rotational artifacts. Hence, direct application of general-purpose MOT models to plant data can lead to poor leaf tracking performance over time, as shown in our experiments. Furthermore, the advancement of deep learning methods for leaf tracking has been limited by the lack of large-scale, high-quality annotated datasets that capture the complexity of real-world agricultural conditions. Existing top-down RGB datasets are typically small in size, span only a short observation period, and have relatively low resolutions, constraining both the development of robust models and the establishment of comparative benchmarks.

In this work, we address these limitations with three main contributions:

(i) **CanolaTrack Dataset**. We present CanolaTrack, the largest high-resolution dataset for leaf tracking in crops to the best of our knowledge. It contains top-down images of 184 canola plants collected over 31 consecutive days. The dataset comprises 5704 RGB images and 31,840 annotated leaf bounding boxes with persistent identity labels. It captures realistic biological events, such as leaf emergence, senescence, occlusion, reappearance, non-uniform growth, and pot rotation.

(ii) **LeafTrackNet.** We propose an efficient tracking framework combining a fine-tuned YOLOv10 detector (Wang et al., 2024) with a MobileNetV3 embedding head (Howard et al., 2019) trained with triplet margin loss. Leaf detections are associated using cosine similarity and Hungarian assignment to maintain identity through growth, occlusion, and reappearance.

(iii) **Cross-dataset Evaluation.** We evaluate LeafTrackNet on the CanolaTrack, KOMATSUNA, and MSU-PID datasets using identical inference hyperparameters without dataset-specific tuning. Under default initialization, LeafTrackNet achieves the highest HOTA on three datasets. Results are reported as mean ± standard deviation over three independent runs.

## 2. Related work

### 2.1. Leaf-level tracking datasets

Despite the growing interest in automated plant phenotyping, leaf tracking remains underexplored compared to leaf classification, segmentation, and counting (Hughes and Salathe, 2016; Wei et al., 2024; Teimouri et al., 2018). As summarized in Table 1, only a few publicly available datasets offer temporal annotations of individual leaves from a top-down view. **LeTra** (Jurado-Ruiz et al., 2024) consists of 513 chlorophyll fluorescence images collected from nine *Arabidopsis thaliana* plants that were photographed three times per day over a 19-day period, resulting in 57 time points and observations of 204 leaves. **KOMATSUNA** (Uchiyama et al., 2017) contains approximately 300 RGB-D images of five *Komatsuna* plants and 900 multiview RGB images of another five plants, recorded every four hours over ten days using both RGB and depth sensors, totaling 1200 RGB images. **MSU-PID** (Cruz et al., 2016) is a multimodal dataset featuring two plant species. The Arabidopsis subset includes 2160 top-view frames per modality from 16 plants, and the bean subset includes 325 frames from five plants. Each frame contains aligned fluorescence, infrared, RGB, and depth images. However, only 576 images from the Arabidopsis subset are annotated. Beyond these top-down datasets, **Pheno-Track3D** (Daviet et al., 2022) captures side-view images of 60 maize hybrids. For each plant, 12 RGB images are taken daily from different angles (30° apart) to enable 3D reconstruction. While this approach enables detailed structural analysis, it requires complex multi-camera setups and image alignment, making it less practical than top-down imaging, in which a single plant image is typically recorded per day.

### 2.2. Plant-specific leaf tracking methods

Several recent approaches have adapted object detection and segmentation pipelines for leaf tracking in plant phenotyping. LeTra (Jurado-Ruiz et al., 2024) leverages a Mask R-CNN backbone for leaf segmentation and applies a heuristic mask-matching strategy based on Intersection-over-Union (IoU) to associate leaves across time frames. While this approach is effective under controlled conditions, LeTra

**Table 1**
Comparison of publicly available datasets for top-down leaf tracking.

| Dataset | Species | Modality | #A. Images | Resolution | #Plants | #Leaves | $\Delta t$ | Rot. |
|---|---|---|---|---|---|---|---|---|
| LeTra | Arabidopsis | F | 0 | 266 × 266 | 9 | 204 | 8 | ✗ |
| KOMATSUNA | Komatsuna | R, D | 1200 | 480 × 480, 166 × 190 | 10 | – | 4 | ✗ |
| MSU-PID | Arabidopsis | F, I, R, D | 576 | ~120 × 120 | 16 | – | 1.6 | ✗ |
| MSU-PID | Bean | F, I, R, D | 172 | 380 × 720 | 5 | – | 1.8 | ✗ |
| **CanolaTrack(Ours)** | Canola | R | 5704 | 1200 × 1200 | 184 | 31,840 | 24 | ✓ |

"#A. Images" = number of annotated images; $\Delta t$ = hours between successive images per plant; "Rot." = pot rotation included; "F", "I", "R", "D" = fluorescence, infrared, RGB, depth; "–" = not reported in original publication.

struggles with occlusion, substantial leaf deformation, and overlapping leaves, which commonly occur during the growth of structurally complex crops such as canola. Plant-Doctor (Montagut-Marquès et al., 2025) integrates YOLOv8 for leaf detection with DeepSORT for identity tracking based on appearance embeddings. Although the incorporation of a re-identification (ReID) module improves association consistency, the embedding model is not specifically trained to capture morphological variations or growth-stage dynamics of leaves, limiting its performance in dense and heterogeneous canopies. Overall, these methods often rely on assumptions such as fixed camera views, minimal leaf overlap, and low temporal variation, which do not hold in realistic long-term scenarios as exemplified by the CanolaTrack dataset.

### *2.3. General multi-object tracking methods*

In the broader computer vision community, multi-object tracking (MOT) has achieved rapid progress, particularly in domains such as pedestrian and vehicle tracking (Yu et al., 2020; Sun et al., 2022). Recent state-of-the-art methods range from tracking-by-detection pipelines to end-to-end models that jointly optimize detection and association. ByteTrack (Zhang et al., 2022) employs an IoU-based association strategy, incorporating both high- and low-confidence detections in a two-stage matching process to improve robustness against occlusions and missed detections. However, its reliance solely on bounding-box geometry limits its applicability in plant scenarios. BoT-SORT (Aharon et al., 2022) enhances traditional tracking-by-detection frameworks by incorporating appearance embeddings, Kalman-filtered motion modeling, and global motion compensation. While more robust than purely geometric methods, BoT-SORT assumes smooth, rigid-body motion, which is frequently violated in leaf tracking due to growth-induced nonlinearity and occlusions. MOTRv2 (Zhang et al., 2023) adopts a transformer-based architecture that jointly performs detection and tracking through query propagation. It achieves strong results on benchmark datasets such as DanceTrack (Sun et al., 2022) and BDD100K (Yu et al., 2020), but requires extensive labeled video data and struggles to generalize to fine-grained, domain-specific tasks such as leaf tracking. MOTIP (Gao et al., 2025) reframes multi-object tracking as an in-context identity prediction problem. It learns association end-to-end by decoding the identity of each current detection while conditioning on historical trajectories and their identity labels. This formulation avoids handcrafted motion or IoU matching rules and achieves strong performance on challenging benchmarks such as DanceTrack.

In summary, most existing MOT methods are developed for high-frame-rate domains with strong frame-to-frame continuity. They assume near-rigid motion, distinctive appearance, and smooth trajectories. These assumptions do not hold in plant imagery. Leaf tracking requires identity persistence across large temporal gaps (e.g., daily frames) with growth-driven appearance drift, frequent self-occlusion, emergence and disappearance events, and occasional global rotations. Under these conditions, simplistic association heuristics (e.g., IoU in LeTra) and general-purpose ReID embeddings (e.g., Plant-Doctor) are unreliable. We therefore introduce **CanolaTrack**, a domain-specific benchmark, and **LeafTrackNet**, a robust tracking framework that combines learned leaf appearance embeddings with a temporal memory for reliable identity maintenance without motion priors.

## 3. Method

The design of LeafTrackNet is guided by failure cases observed when applying MOT trackers to biologically complex plant growth sequences. A central design choice is to perform identity association using invariant features extracted from cropped and resized leaf images. This avoids primary reliance on geometric cues such as centroid position, leaf overlap, or leaf size. First, geometric association methods, such as IoU or Kalman-based tracking, often fail under occlusions and pose changes, which are frequent due to overlapping leaves and rotational artifacts. Second, generic embedding extractors often lack the discriminative capacity to distinguish visually similar leaf instances within a single plant. Finally, end-to-end transformers designed for high-frame-rate pedestrian tracking can fail to generalize in temporally sparse, biologically dynamic sequences. These observations motivate a framework that decouples spatial localization from identity matching and learns representations robust to temporal sparsity and discontinuities (Fig. 1). During inference, cosine similarity and memory-based matching support identity propagation without relying on geometric continuity.

### *3.1. Training phase*

***Triplet sampling***. Let $I_k^t \in \mathbb{R}^{3\times W_r\times H_r}$ denote the raw RGB image of plant $k$ at time $t$ with spatial resolution $W_r \times H_r$. The corresponding set of annotated leaves in an image is defined as

$$\mathcal{G}_k^t = \left\{b_{k,i}^t\right\}_{i=1}^{L_k^t} = \left\{b_{k,1}^t, \dots, b_{k,L_k^t}^t\right\},$$

where each element $b_{k,i}^t = (u_{k,i}^t, v_{k,i}^t, w_{k,i}^t, h_{k,i}^t)$ represents a ground-truth bounding box of leaf $i$. The coordinates $u_{k,i}^t$ and $v_{k,i}^t$ specify the top-left corner, while $w_{k,i}^t$ and $h_{k,i}^t$ denote the width and height, respectively.

To obtain an individual leaf representation, we first crop the leaf region defined by each bounding box and resize it to a fixed resolution. To this end, we define a crop-and-resize operator $\psi : \mathbb{R}^{3\times W_r\times H_r} \times \mathbb{R}^4 \to \mathbb{R}^{3\times W\times H}$, which extracts the leaf region specified by bounding box $b$ from image $I$ and resizes it to a fixed spatial resolution $(W, H)$. The resulting resized leaf crop of leaf $i$ of plant $k$ at time $t$ is given by

$$x_{k,i}^t = \psi\left(I_k^t, b_{k,i}^t\right) \in \mathbb{R}^{3\times W\times H}. \tag{1}$$

For robust identity tracking across time, the embedding network must learn to distinguish observations of the same leaf across time from observations of different leaves. Therefore, we employ a triplet loss function, which guides the model using sets of three samples: an anchor, a positive, and a negative. The embeddings of the same leaf at different time points are pulled closer together, while embeddings of different leaves are pushed further apart in the feature space. To utilize this contrastive training strategy, we construct triplets $(x_a, x_p, x_n)$. For each anchor $x_a$, randomly selected from a plant at a given time point, the positive and negative samples are chosen as follows.

(i) *Positive Selection:* Given an anchor crop $x_a = x_{k,i}^{t_a}$, the positive sample is a crop of the same plant at a different randomly selected time point,

$$x_p = x_{k,i}^{t_p} \quad s.t.\ t_p \neq t_a.$$

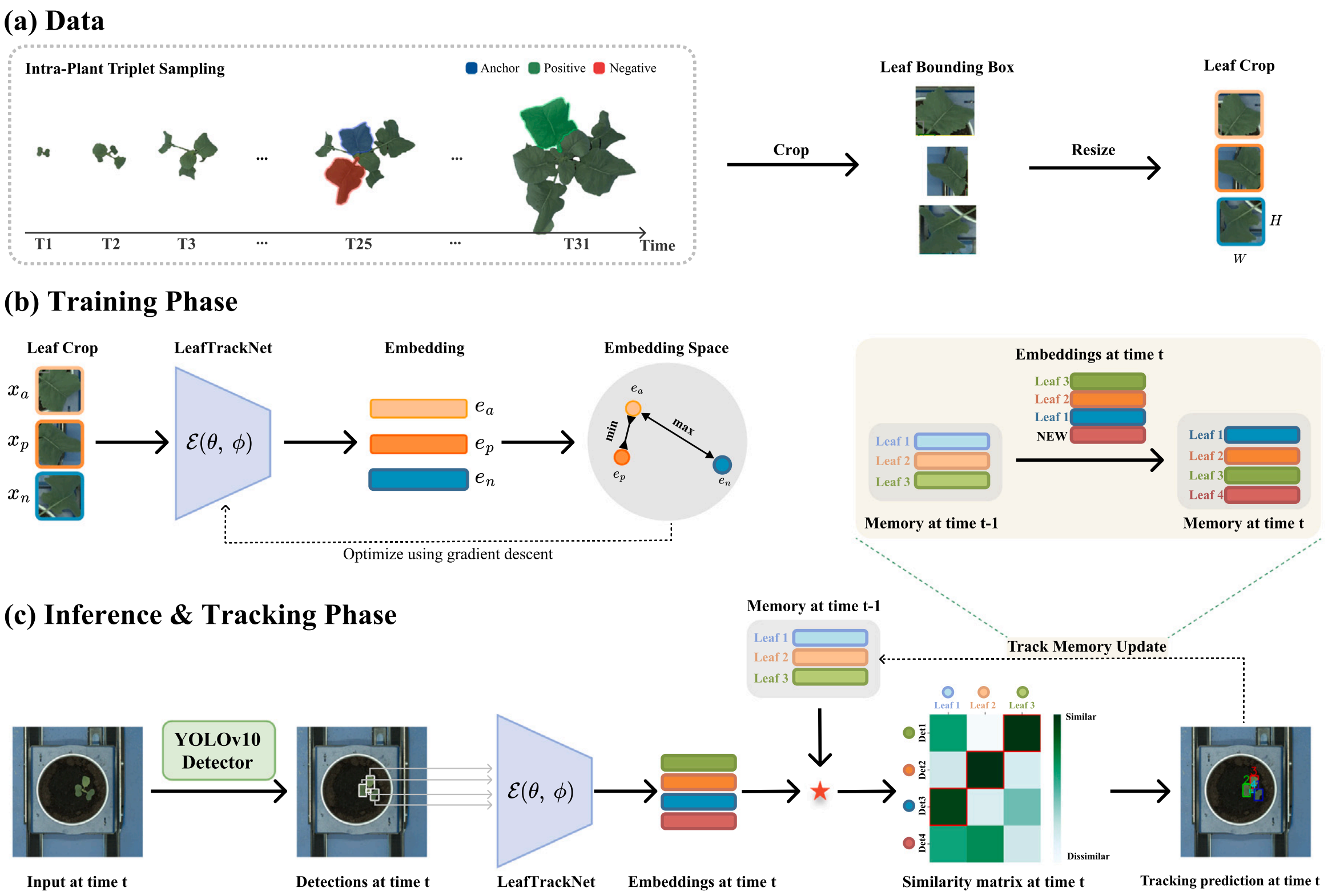

**Fig. 1.** LeafTrackNet overview. (a) Data: intra-plant triplet sampling and leaf-crop extraction. (b) Training Phase: LeafTrackNet processes anchor, positive, and negative leaf crops and is trained using triplet margin loss to learn a discriminative and temporally consistent embedding space. (c) Inference & Tracking Phase: YOLOv10-based leaf detection is followed by matching against the memory bank using cosine-similarity and Hungarian assignment. The ⋆ symbol denotes cosine similarity.

(ii) *Negative Selection:* Given an anchor crop $x_a = x_{k,i}^{t_a}$, the negative sample is a crop of a different leaf from the same plant at a randomly chosen time point

$$x_n = x_{k,j}^{t_n} \quad s.t.\ j \neq i.$$

(iii) *Triplet Formation:* The final training triplet consists of two crops from the same leaf $(x_a, x_p)$ and one crop from a different leaf $(x_n)$:

$$(x_a, x_p, x_n) = \left(x_{k,i}^{t_a}, x_{k,i}^{t_p}, x_{k,j}^{t_n}\right). \tag{2}$$

***Model architecture.*** The LeafTrackNet model $\mathcal{E} : \mathbb{R}^{3\times W\times H} \to \mathbb{R}^D$ is defined as:

$$\mathcal{E}(x) = \mathcal{F}_\phi\big(\mathcal{N}_\theta(x)\big) \tag{3}$$

where $\mathcal{N}_\theta : \mathbb{R}^{3\times W\times H} \to \mathbb{R}^F, F \in \mathbb{N}$ is a MobileNetV3 backbone pretrained on ImageNet (Deng et al., 2009) and truncated before its classification head, and $\mathcal{F}_\phi : \mathbb{R}^F \to \mathbb{R}^D, D \in \mathbb{N}$ is a linear projection layer mapping the feature vector to the embedding space, with parameters $(\theta, \phi)$ respectively.

***Loss function.*** Given a triplet $(x_a, x_p, x_n)$ as defined in Eq. (2), the corresponding embeddings are

$$e_a = \mathcal{E}(x_a), \quad e_p = \mathcal{E}(x_p), \quad e_n = \mathcal{E}(x_n). \tag{4}$$

We adopt the triplet margin loss as introduced in Vassileios Balntas and Mikolajczyk (2016) to enforce that the anchor–positive distance is smaller than the anchor–negative distance by a margin $m$:

$$\mathcal{L}(x_a, x_p, x_n) = \max\left\{0,\ \|e_a - e_p\|_2^2 - \|e_a - e_n\|_2^2 + m\right\}, \tag{5}$$

such that the model is trained to learn an embedding space in which leaves of the same identity are close and those of different identities are well separated.

***Training details.*** The leaf detector uses YOLOv10x initialized with the official COCO-pretrained weights. It is fine-tuned on the CanolaTrack training set for 100 epochs using 640 × 640 inputs and a batch size of 48 across four NVIDIA Tesla V100S GPUs. Optimization uses AdamW with an initial learning rate of $2 \times 10^{-3}$ and a weight decay of $5 \times 10^{-4}$.

The embedding network uses an ImageNet-pretrained MobileNetV3-Large backbone and a 128-dimensional embedding layer. Its parameters $(\theta, \phi)$ are optimized using the Adam optimizer with an initial learning rate of $1 \times 10^{-4}$ and a weight decay of $1 \times 10^{-5}$. Following the setup in Zhu et al. (2020), the triplet loss margin is set to $m = 0.3$. Following Howard et al. (2019), all leaf crops are resized to 224 × 224. Training is conducted for up to 80 epochs with a batch size of 48 on four NVIDIA Tesla V100S GPUs, using an early stopping strategy.

### 3.2. Inference and tracking phase

In contrast to the training phase, where triplets are sampled randomly across time, the inference and tracking phase proceeds sequentially, processing images in temporal order from the first to the last observation. Inspired by MOTRv2 (Zhang et al., 2023), we use YOLOv10 (Wang et al., 2024) as our leaf detector, fine-tuned on the CanolaTrack training set. To reduce false positives, we filter out leaf detections with confidence scores below 0.5. For each plant image $I_k^t$, the YOLOv10 detector provides a set of leaf bounding boxes:

$$\mathcal{D}^t = \left\{\hat{b}_i^t\right\}_{i=1}^{L} = \left\{\hat{b}_1^t, \dots, \hat{b}_L^t\right\},$$

where each $\hat{b}_i^t = (\hat{u}_i^t, \hat{v}_i^t, \hat{w}_i^t, \hat{h}_i^t)$ denotes the top-left coordinates, width, and height of the $i$th leaf bounding box at time $t$. Each detected region is then cropped and resized using the operator $\psi$ (as in Eq. (1)) and subsequently encoded into an embedding using the trained network $\mathcal{E}$:

$$\hat{e}_i^t = \mathcal{E}\big(\psi(I_k^t, \hat{b}_i^t)\big). \tag{6}$$

***Tracking memory bank.*** To improve long-term leaf tracking, we introduce a tracking memory bank $\mathcal{T}^t$ containing the active tracks at time $t$. Each track represents the identity assigned to one tracked leaf. Each tracked leaf $\ell$ is represented by a prototype embedding vector $p_\ell^t \in \mathbb{R}^D$ and an age counter $a_\ell^t \in \mathbb{N}$, which records the number of consecutive images in which the leaf is not present:

$$\mathcal{T}^t = \left\{(p_\ell^t, a_\ell^t)\right\}_{\ell=1}^{N_t},$$

where $N_t \in \mathbb{N}$ is the number of active tracks at time $t$.

***Initialization at*** $t = 1$. Since no tracks exist initially, all detections within the first image are treated as new tracks. Let $\{e_j^1\}_{j=1}^{N_1}$ denote the embeddings extracted from the detections at time $t = 1$. The memory bank is initialized as

$$\mathcal{T}^1 = \left\{(p_j^1 = e_j^1,\ a_j^1 = 0)\right\}_{j=1}^{N_1},$$

where $N_1$ is the number of detected leaves in the first image.

***Sequential update at*** $t > 1$. At each subsequent time $t$, the current set of leaf embeddings $\{e_j^t\}_{j=1}^{N_t}$ is matched to the set of existing prototype embeddings from the previous image $\{p_\ell^{t-1}\}_{\ell=1}^{N_{t-1}}$ stored in the memory bank. To match the embeddings, we compute the following similarity matrix $S$ using cosine similarity:

$$S_{\ell j} = \left(p_\ell^{t-1}\right)^\top e_j^t \quad \in \mathbb{R}^{N_{t-1}\times N_t}. \tag{7}$$

Afterwards, to determine the optimal one-to-one assignment between current detections and the existing tracks, we apply the Hungarian algorithm (Kuhn, 1955). The similarity matrix $S$ in Eq. (7) contains the similarities between $N_{t-1}$ tracks and $N_t$ detections at time $t$. We then transform it into a cost matrix $C = 1 - S_{\ell j} \in \mathbb{R}^{N_{t-1}\times N_t}$, where a lower cost corresponds to a higher similarity.

We then seek one-to-one assignments $\pi : \{1,\dots,N_{t-1}\} \to \{1,\dots,N_t\}$, where $\pi(\ell) = j$ indicates assigning track $\ell$ to detection $j$. The Hungarian algorithm is used to find the optimal assignment $\pi^*$ by minimizing the total cost, i.e., maximizing the total similarity, in $\mathcal{O}\left(\max(N_{t-1}, N_t)^3\right)$ time:

$$\pi^* = \arg\min_\pi \sum_{\ell=1}^{N_{t-1}} C_{\ell,\pi(\ell)} = \arg\min_\pi \sum_{\ell=1}^{N_{t-1}} \left(1 - (p_\ell^{t-1})^\top e_{\pi(\ell)}^t\right) \tag{8}$$

Tracks $\ell$ without a valid assignment or with $S_{\ell,\pi^*(\ell)} < \tau_s$, where $\tau_s$ denotes the similarity threshold, are treated as unmatched. The memory bank is then updated as follows:

(i) *Matched Detections (Persistent or Reappearing Leaves):* For each matched pair $(\ell, j)$, the track's prototype is updated by an exponential moving average:

$$p_\ell^t = (1-\alpha)\cdot p_\ell^{t-1} + \alpha \cdot e_j^t, \quad a_\ell^t = 0,$$

where $\alpha \in [0,1]$ controls temporal smoothing.

(ii) *Unmatched Detections (New Leaves):* Each unmatched detection $j$ is initialized as a new track in the memory bank:

$$p_{\ell'}^t = e_j^t, \quad a_{\ell'}^t = 0,$$

where $\ell' = N_{t-1}+1,\ N_{t-1}+2,\ \cdots$.

(iii) *Unmatched Tracks (Disappearing Leaves):* Existing tracks in the memory bank that are not matched to the current set of leaves are considered temporarily or permanently absent. For any unmatched track $\ell$, we keep the embedding prototype unchanged and increment its age:

$$p_\ell^t = p_\ell^{t-1}, \quad a_\ell^t = a_\ell^{t-1} + 1.$$

Any tracked leaf whose age exceeds the threshold ($a_\ell^t > \tau_a$) is removed from the memory bank.

***Inference details.*** Leaf detections with confidence scores below 0.5 are discarded. We use a similarity threshold of $\tau_s = 0.4$, an age threshold of $\tau_a = 5$, and a smoothing coefficient of $\alpha = 0.5$ during inference for all datasets. All inference experiments were conducted on a single NVIDIA Tesla V100S GPU.

## 4. Experiments and results

### 4.1. Dataset

We constructed a large-scale leaf tracking dataset, **CanolaTrack**, by continuously capturing top-down RGB images of 184 canola plants over 31 days during their early growth cycle. The dataset comprises 5704 images at a resolution of $1200 \times 1200$. Images from each plant sequence were annotated chronologically using LabelMe[1] by manually drawing tight, axis-aligned bounding boxes around the visible extent of each leaf, producing 31,840 annotated leaf instances with persistent identities maintained across frames. Fully occluded or absent leaves received no bounding box, and their identities were restored upon reappearance. The annotations were stored in MOT format as (frame, leaf ID, $x, y, w, h$). As illustrated in Fig. 2, image acquisition began at the emergence of the first leaves and continued until the formation of floral buds, covering an entire growth period characterized by dynamic leaf expansion. Within a single plant, leaves are highly similar in color and texture, while simultaneously exhibiting strong intra-leaf variability over time due to growth, deformation, partial visibility, and orientation changes. CanolaTrack further increases identity ambiguity by mixing newly emerged small leaves with mature large leaves in the same frame.

To highlight the scale and characteristics of CanolaTrack, we provide a comparative overview of existing top-down leaf tracking datasets in Table 1, including LeTra (Jurado-Ruiz et al., 2024), KOMATSUNA (Uchiyama et al., 2017), and MSU-PID (Cruz et al., 2016), focusing on key factors such as species, number of annotated RGB images, image resolution, number of plants and leaf instances, and the presence of pot rotation.

We characterize the distributional properties of each tracking dataset in Fig. 3(a). We exclude LeTra from this analysis because it does not provide per-leaf identity annotations for tracking (only object detection labels). For each tracked leaf identity, we quantify temporal variability by computing empirical distributions of (i) relative scale change between successive frames (bounding-box area ratio), (ii) the intersection-over-union (IoU) between bounding boxes in adjacent frames, and (iii) centroid displacement normalized by leaf area. The long-tail regimes (highlighted in pink) correspond to the most challenging conditions, including abrupt scale changes between adjacent frames (up to ∼4× increase in area), near-zero spatial overlap for the same leaf across frames (IoU approaching 0), and large centroid shifts (up to ∼3× in normalized displacement). A higher proportion of such long-tail cases, together with a median (highlighted in orange) shifted toward the tail, indicates that CanolaTrack exhibits stronger growth and motion-induced variability than KOMATSUNA and MSU-PID. We additionally provide representative images and the corresponding per-leaf motion trajectories in Fig. 3(b). Komatsuna leaves in KOMATSUNA and bean leaves in MSU-PID form compact clusters in the geometric trajectory space, reflecting relatively smooth inter-frame dynamics. In contrast, canola leaves in CanolaTrack present greater challenges due to global plant rotations and sparser temporal sampling, which jointly induce larger point-to-point displacements and more irregular per-leaf trajectories.

For all subsequent experiments, we randomly partition the CanolaTrack dataset into training and test sets following a standard 80/20 split. The training set consists of 147 plants (4557 images, 25,485 leaves), and the test set contains 37 previously unseen plants, with 1147 images and 6355 leaves. We apply the same 80/20 protocol to KOMATSUNA and MSU-PID. The detailed statistics are reported in Table 2. All benchmarked models are trained exclusively on the training set and evaluated on the test set for fair and consistent comparison.

[1] https://github.com/wkentaro/labelme.

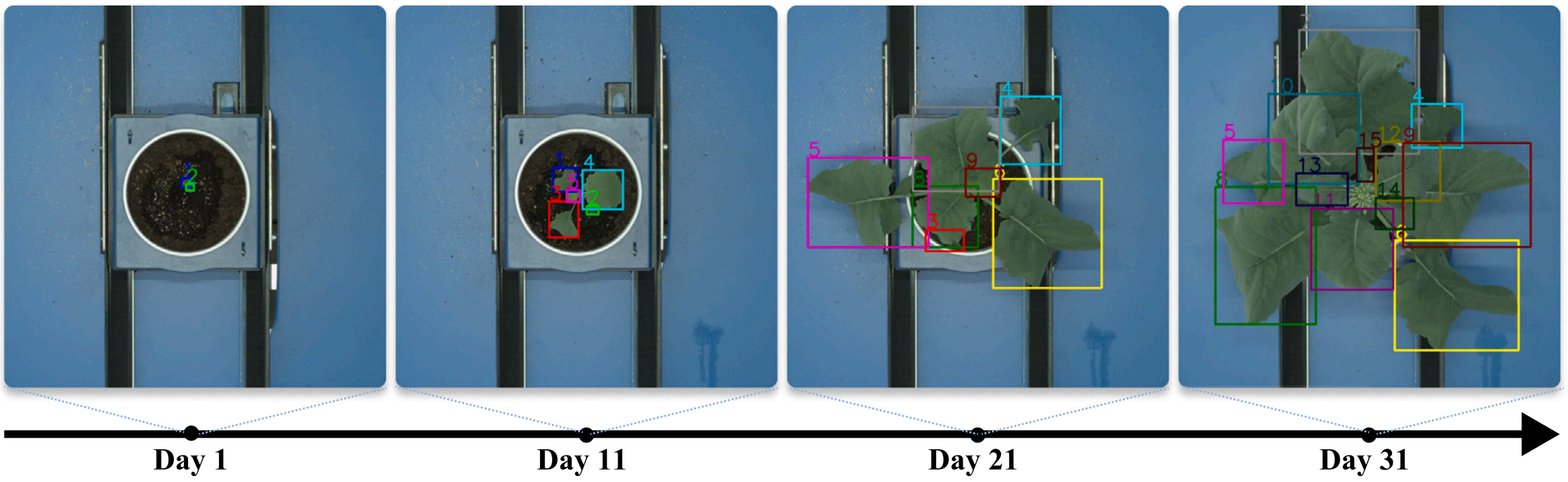

**Fig. 2.** Example RGB images of Plant-003 on days 1, 11, 21, and 31, with color-coded bounding boxes indicating individual leaves over time.

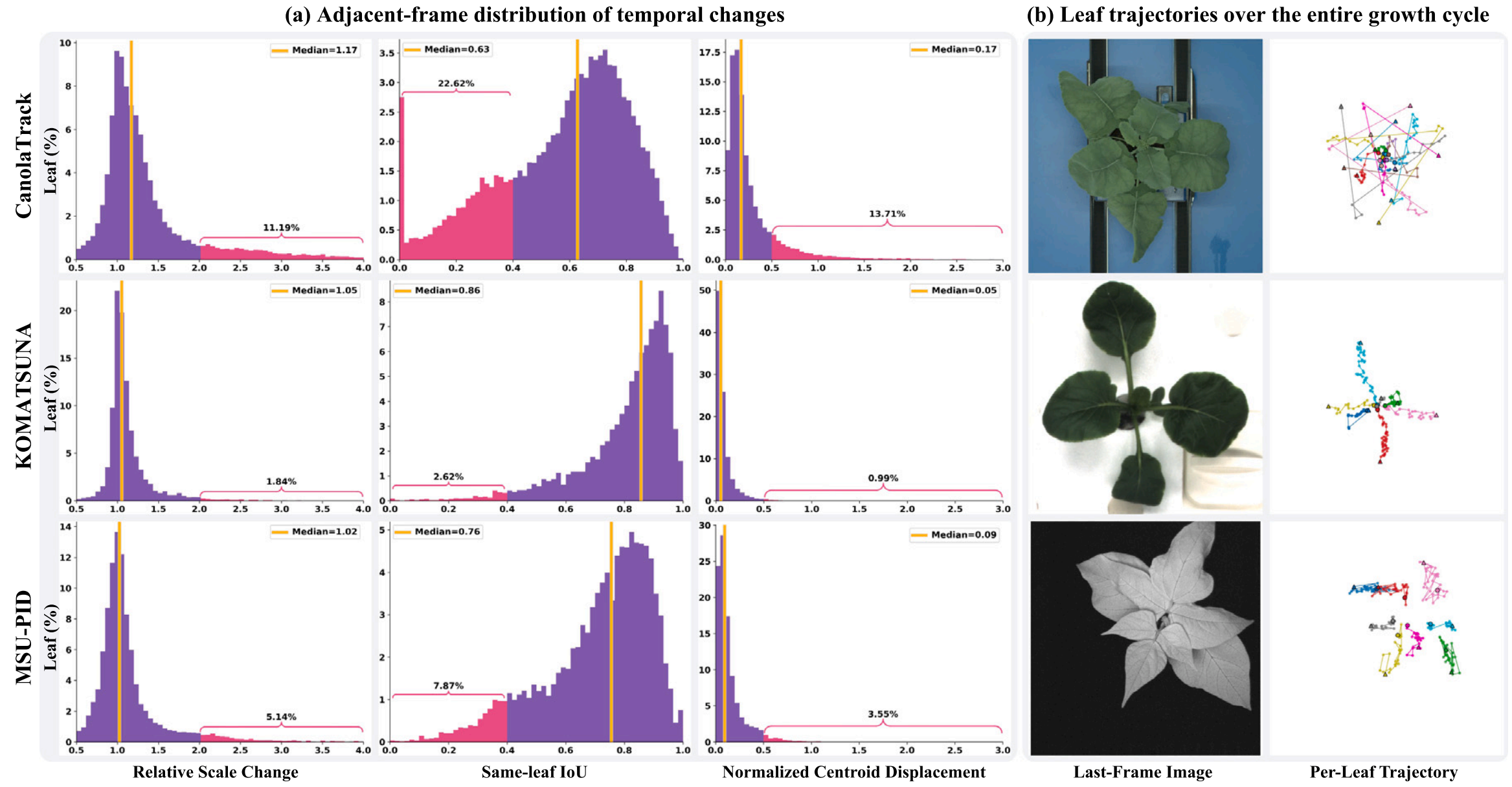

**Fig. 3.** Frame-to-frame leaf dynamics in CanolaTrack, KOMATSUNA, and MSU-PID. (a) Distributions of relative scale change, same-leaf intersection over union (IoU), and leaf-area-normalized centroid displacement between adjacent frames. Pink shading indicates challenging long-tail cases and their proportions, and orange lines indicate the medians. (b) Final-frame examples with per-leaf trajectories over the full growth cycle. Each leaf is represented by a distinct color, with circles and triangles marking the start and end of each trajectory, respectively.

### 4.2. Evaluation metrics

Following common practice in multi-object tracking evaluation, we use Higher Order Tracking Accuracy (HOTA) as the primary metric for evaluating tracking performance (Zhang et al., 2023; Gao et al., 2025; Jurado-Ruiz et al., 2024; Luiten et al., 2021). We also report additional MOT metrics, including Detection Accuracy (DetA), Association Accuracy (AssA), Multi-Object Tracking Accuracy (MOTA), Identification F1 Score (IDF1), the number of identity switches (IDSW), together with inference speed in frames per second (FPS).

**HOTA** measures the joint performance of detection DetA and association AssA, providing a balanced evaluation of tracking quality

$$\mathrm{HOTA} = \sqrt{\mathrm{DetA} \times \mathrm{AssA}}. \tag{9}$$

**Table 2**
Summary of the dataset partitions used in our benchmarking experiments.

| Dataset | Modality | Split | Seq | Frames | Leaves |
|---|---|---|---|---|---|
| KOMATSUNA | R | train | 16 | 960 | 4663 |
| | | test | 4 | 240 | 1163 |
| MSU-PID | F, I, R | train | 45 | 1614 | 13920 |
| | | test | 13 | 464 | 4162 |
| CanolaTrack | R | train | 147 | 4557 | 25485 |
| | | test | 37 | 1147 | 6355 |

Since each plant has multiple camera views in KOMATSUNA and multiple modalities in MSU-PID, the number of sequences reported here exceeds the number of plants in Table 1.

**DetA** quantifies how well the tracker detects objects across images. Let TP, FP, and FN represent true positives, false positives, and false negatives, respectively. DetA is computed as

$$\mathrm{DetA} = \frac{\mathrm{TP}}{\mathrm{TP} + \mathrm{FP} + \mathrm{FN}}. \tag{10}$$

**Table 3**
Tracking performance on the CanolaTrack dataset. Best scores are in **bold**; second-best scores are underlined. Results are reported as mean ± standard deviation across three runs.

| Domain | Method | HOTA↑ | DetA↑ | AssA↑ | MOTA↑ | IDF1↑ | IDSW↓ | FPS↑ |
|---|---|---|---|---|---|---|---|---|
| General | BoT-SORT | 57.05±0.14 | 66.77±0.06 | 49.03±0.29 | 77.28±0.10 | 69.24±0.23 | 274±4 | 8.75 |
| | ByteTrack | 52.49±0.27 | 59.92±0.14 | 46.58±0.36 | 73.23±0.36 | 68.99±0.43 | 282±4 | 9.04 |
| | MOTRv2 | 78.30±1.36 | 77.33±1.65 | 79.36±1.03 | 79.68±1.75 | 83.78±1.39 | **67**±7 | 15.79 |
| | MOTIP | 79.68±0.25 | 88.83±0.02 | 71.61±0.99 | 88.00±0.14 | 81.03±0.54 | 460±39 | 15.91 |
| Plant | LeTra* | 67.02±0.20 | 82.03±0.37 | 54.98±0.40 | 82.09±0.44 | 69.06±0.32 | 547±3 | 23.41 |
| | Plant-Doctor | 58.94±0.11 | 72.31±0.17 | 48.36±0.30 | 77.76±0.09 | 69.59±0.20 | 821±6 | 23.69 |
| | **LeafTrackNet** | **88.03**±0.24 | **92.25**±0.03 | **84.07**±0.49 | **93.64**±0.18 | **92.90**±0.35 | 245±15 | 24.64 |
| | LeafTrackNet† | 56.10±0.35 | 88.54±0.22 | 35.64±0.40 | 47.59±0.50 | 47.47±0.38 | 3022±10 | **24.96** |
| | LeafTrackNet‡ | 85.54±0.43 | 88.49±0.21 | 82.80±0.64 | 92.73±0.39 | 92.07±0.39 | 149±16 | 24.72 |

LeTra* is adapted from mask-based to box-based matching for CanolaTrack. LeafTrackNet† and LeafTrackNet‡ denote embedding-only greedy matching with age thresholds of 0 and 5, respectively.

**AssA** measures the correctness of identity preservation over time. Let T denote the set of time steps, and let $\mathrm{TP}^t_{\mathrm{assoc}}$ denote the number of correctly associated detections at time $t$. AssA is defined as the average fraction of correctly associated objects given that a detection is matched:

$$\mathrm{AssA} = \frac{1}{|\mathrm{T}|} \sum_{t \in \mathrm{T}} \frac{\mathrm{TP}^t_{\mathrm{assoc}}}{\mathrm{TP}^t_{\mathrm{assoc}} + \mathrm{FP}^t_{\mathrm{assoc}} + \mathrm{FN}^t_{\mathrm{assoc}}}. \tag{11}$$

**MOTA** considers missed detections, false positives, and identity switches. An identity switch occurs when a ground-truth object is matched to a different predicted identity than in its previous match. Let $\mathrm{GT}_t$ denote the number of ground-truth objects at time $t$ and let $\mathrm{IDSW}_t$ denote the number of identity switches. MOTA and the total number of identity switches are defined as

$$\mathrm{MOTA} = 1 - \frac{\sum_t (\mathrm{FN}_t + \mathrm{FP}_t + \mathrm{IDSW}_t)}{\sum_t \mathrm{GT}_t}, \qquad \mathrm{IDSW} = \sum_t \mathrm{IDSW}_t. \tag{12}$$

**IDF1** is the F1 score of correctly identified detections. Here, IDTP, IDFP, and IDFN denote identity-level true positives, false positives, and false negatives, respectively.

$$\mathrm{IDF1} = \frac{2 \cdot \mathrm{IDTP}}{2 \cdot \mathrm{IDTP} + \mathrm{IDFP} + \mathrm{IDFN}}. \tag{13}$$

**FPS** measures inference efficiency by reporting the number of frames processed per second:

$$\mathrm{FPS} = \frac{N_{\mathrm{f}}}{T}, \tag{14}$$

where $N_{\mathrm{f}}$ denotes the total number of processed frames and $T$ the corresponding wall-clock runtime in seconds.

### 4.3. Experiment results

In the following, we evaluate our proposed method on the CanolaTrack, KOMATSUNA and MSU-PID datasets and compare it against state-of-the-art approaches using the TrackEval tool.[2] We compare LeafTrackNet with general-purpose trackers, including BoT-SORT, ByteTrack, MOTRv2, and MOTIP, and plant-specific trackers, including LeTra and Plant-Doctor. The baseline models are trained using their original implementations and the default parameter settings provided by the authors.

#### 4.3.1. CanolaTrack

**Quantitative Benchmarking.** Table 3 reports a performance comparison between our proposed LeafTrackNet and state-of-the-art methods across six standard MOT metrics and an inference-speed metric. A key observation is that LeafTrackNet's gains are not driven by detection alone. While it also provides the best DetA (92.25), the largest separation from competitive baselines appears in association-related metrics. Compared with the general tracker with the highest HOTA, MOTIP, LeafTrackNet improves HOTA by 8.35 points, AssA by 12.46, and IDF1 by 11.87 and reduces IDSW from 460 to 245.

Table 3 reveals that common MOT pipelines underperform in this domain. BoT-SORT and ByteTrack exhibit low AssA (49.03 and 46.58), consistent with association failures when motion smoothness assumptions break (growth-induced displacement, leaf emergence and disappearance, frequent overlap and global rotation). Transformer-based trackers (MOTRv2, MOTIP) improve general performance (≈78–80 HOTA), yet their identity preservation still lags behind LeafTrackNet (≤83.78 IDF1), indicating that plant scenes contain weaker instance-specific appearance cues than pedestrian/vehicle benchmarks and require stronger leaf-centric embeddings and geometric and temporal constraints. MOTRv2 records the fewest IDSW at 67, compared with 245 for LeafTrackNet. However, this result should be interpreted together with its lower DetA and MOTA of 77.33 and 79.68, compared with 92.25 and 93.64 for LeafTrackNet. Missed or prematurely terminated tracks can also reduce the opportunities for registering identity switches, meaning that a lower IDSW count does not necessarily represent better overall tracking. Moreover, IDSW measures local identity changes, whereas HOTA balances detection accuracy with association quality over complete trajectories (Luiten et al., 2021). Among the plant-specific baselines, LeTra attains competitive detection accuracy (DetA = 82.03) but substantially weaker identity association (AssA = 54.98; IDF1 = 69.06). This discrepancy is consistent with LeTra's reliance on IoU-based matching between masks in consecutive frames, which becomes fragile in dense rosettes where self-occlusion reduces the reliability of purely overlap-driven identity propagation. Overall, the results show that CanolaTrack rewards methods that jointly optimize detection and identity association using plant-specific inductive biases. LeafTrackNet's consistent improvements in HOTA, AssA, and IDF1 underscore its suitability for downstream organ-level phenotyping, for which temporal continuity of individual leaves is critical.

The effectiveness of LeafTrackNet was further investigated by comparing against two simple baselines that rely solely on embedding similarity. LeafTrackNet†, which uses greedy cosine matching with an age threshold of 0, drops to 56.10 HOTA, showing that strict frame-to-frame embedding matching is insufficient when leaves are temporarily occluded or missing. In contrast, LeafTrackNet‡, which keeps the original age threshold of 5, reaches 85.54 HOTA. The full LeafTrackNet further improves HOTA to 88.03, indicating that disappearance-tolerant memory, EMA prototype updating, and Hungarian assignment provide additional gains beyond embedding quality alone.

**Qualitative Comparisons.** Fig. 4 visualizes two main challenges in top-down leaf tracking, namely heavy occlusion and pot rotation.

Fig. 4(a) illustrates an occlusion case during the early stage of canola plant growth, when leaves are small (Day 11 → Day 12). While detecting small leaves is already challenging, the additional complexity introduced by densely overlapping leaves further complicates identity tracking. This is evident from the predictions of BoT-SORT and ByteTrack, which exhibit ID switches due to their reliance on Kalman-filter-based motion models and IoU-based association. These methods assume consistent motion and non-overlapping objects, making them unreliable when leaves overlap or shift unpredictably, as shown by leaves 2, 3, and 4. LeTra, which relies on IoU mask matching, similarly fails under occlusion because overlapping leaves (e.g., leaves 1, 4, and 5) often merge into a single region, leading to segmentation errors and track loss for Leaf 4. Plant-Doctor uses ReID features that were not trained specifically for this task. Occlusion further reduces visible cues such as texture and shape, leading to unstable identity embeddings and ID swaps between leaves 1 and 5. In contrast, our method incorporates both appearance and temporal cues in a leaf-aware representation that remains robust under occlusion, scale variation, and partial visibility and enables consistent leaf tracking, even after temporary occlusion (e.g., Leaves 1, 2, and 4). Notably, our model achieves high performance without relying on computationally intensive transformer-based architectures, such as those used in MOTRv2 and MOTIP. This balance of efficiency and robustness makes our approach especially suited for fine-grained leaf tracking tasks.

[2] https://github.com/JonathonLuiten/TrackEval.

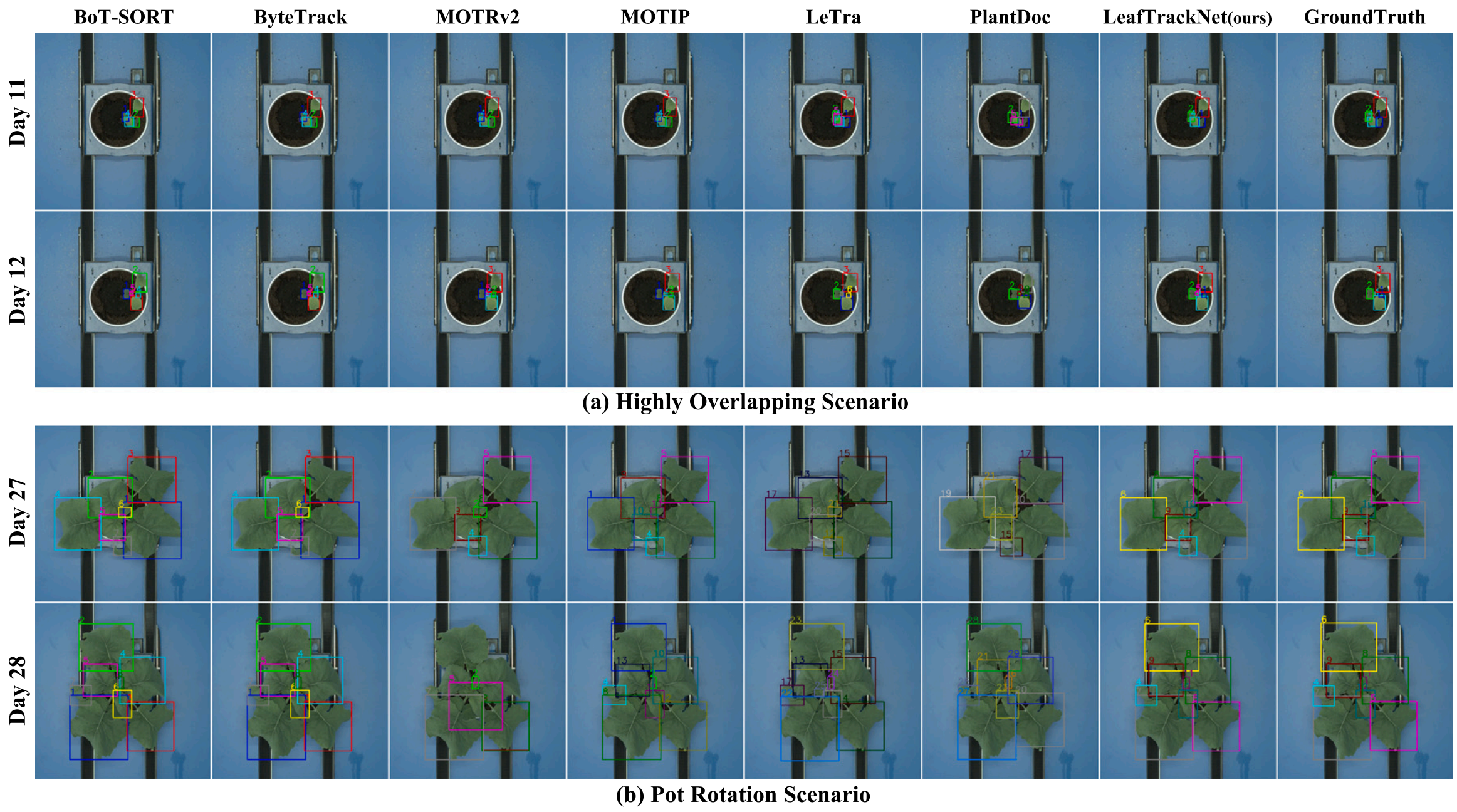

**Fig. 4.** Qualitative tracking results on Plant-158 from the CanolaTrack dataset. (a) High-overlap scenario between Day 11 and Day 12. (b) Pot rotation scenario between Day 27 and Day 28. The ground-truth column is included for reference.

Fig. 4(b) illustrates a rotation scenario in which the entire pot rotated around 90° clockwise from Day 27 to Day 28. This global transformation disrupts spatial continuity and presents a major challenge to most baseline tracking methods. Methods such as BoT-SORT and ByteTrack assume smooth, linear motion and local consistency in object positions via Kalman filtering and IoU-based matching. As a result, they fail to maintain leaf identities under pot rotation as their models assume linear trajectories. LeTra, which also depends on spatial mask overlap, fails similarly. Its performance is further degraded by accumulated errors from previous occlusions. Plant-Doctor uses ReID features that were not trained specifically for this task and are sensitive to viewpoint and orientation changes. Consequently, it treats rotated leaves as new objects, producing frequent identity switches. MOTRv2, which uses learned track queries, fails when those queries no longer align with the spatial positions of rotated proposals, resulting in missed detections. MOTIP predicts the ID of each current detection by decoding it from the detection embedding while attending to embeddings of historical tracklets augmented with ID embeddings. The feature drift due to a sudden rotation can prevent the ID decoder from aligning current detections, leading to fragmented tracks for leaves 5 and 9. In contrast, our method leverages a memory-based embedding strategy in which the embeddings capture both appearance and structural cues that are robust to rotation and geometric transformations. As a result, it successfully re-associates the rotated leaves, demonstrating strong invariance to viewpoint and orientation changes during leaf identity tracking.

**Embedding Stability and Temporal Robustness Analysis.** To further investigate the strengths of our method, we conduct a detailed case study of Plant-158, which exhibits several challenges, including leaf overlap, emergence, deformation, and pot rotation.

Fig. 5(a) shows the leaf area trajectories for 13 individual leaves from a single plant, revealing large variation in growth dynamics: some leaves expand gradually, while others grow rapidly and reach substantially larger sizes. By the fourth week, the leaf area distribution becomes highly variable due to the coexistence of newly emerged small leaves and fully expanded mature ones, creating significant challenges for tracking—particularly for the smaller leaves, which are often occluded or visually similar to nearby structures. Fig. 5(b) shows a t-SNE (van der Maaten and Hinton, 2008) projection of all detected leaf embeddings over the full 31-day life cycle. Each marker corresponds to a leaf instance, and colors denote leaf identities. The visualization reveals that embeddings from the same leaf identity form tight, distinguishable clusters, despite natural variations in morphology, occlusion, and orientation over time. This demonstrates the discriminative power and temporal stability of our learned embedding space. Handling such invariances is essential for leaf tracking in real-world plant phenotyping, where leaf size, shape, orientation, and appearance evolve nonlinearly across time due to plant growth. In contrast, LeafTrackNet learns leaf-specific, invariant feature representations that abstract away from low-level geometric variations and temporal discontinuities. By capturing structural and visual cues that remain stable across time,

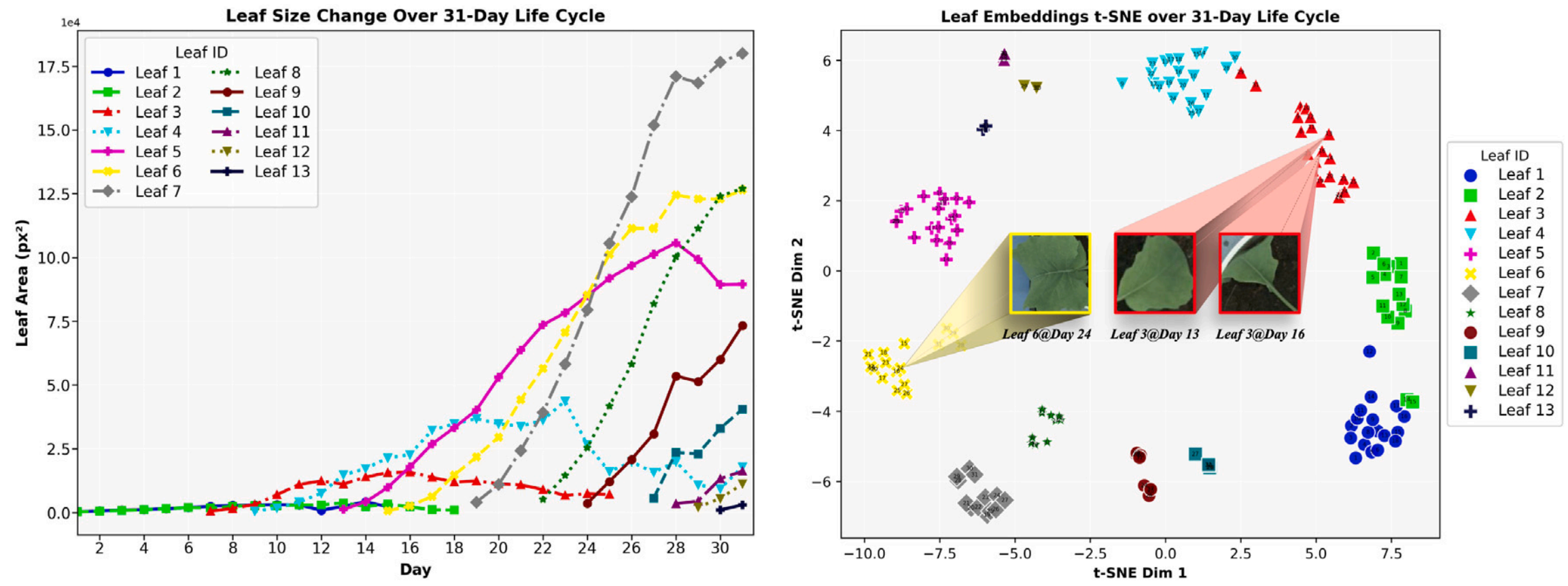

(a) (b)

**Fig. 5.** Leaf dynamics and embedding separability for Plant-158 over the 31-day period. (a) Per-leaf bounding-box area trajectories illustrating emergence, disappearance, and growth. (b) t-SNE projection of learned leaf embeddings with day indices, showing compact same-leaf clusters and separation between different leaves.

**Table 4**
Tracking performance on the KOMATSUNA test set with direct KOMATSUNA fine-tuning (upper block) and sequential CanolaTrack-to-KOMATSUNA fine-tuning (lower block). Values are reported as mean ± standard deviation over three runs.

| Pretraining | Method | HOTA↑ | DetA↑ | AssA↑ | MOTA↑ | IDF1↑ | IDSW↓ |
|---|---|---|---|---|---|---|---|
| Default | BoT-SORT | 82.25±0.30 | 83.75±0.36 | 80.88±0.30 | 93.69±0.56 | 92.93±0.38 | 7±2 |
| | ByteTrack | 76.63±0.22 | 77.45±0.11 | 75.99±0.44 | 92.03±0.30 | 92.42±0.32 | 7±1 |
| | MOTRv2 | 82.39±0.73 | 80.55±1.08 | 84.38±0.54 | 86.62±1.34 | 92.69±0.41 | **3**±1 |
| | MOTIP | 60.39±1.45 | 83.67±0.58 | 43.74±1.85 | 59.92±2.29 | 55.22±2.56 | 280±20 |
| | LeTra | 74.71±0.80 | 74.24±0.48 | 75.33±1.11 | 82.52±0.43 | 85.76±0.74 | 15±2 |
| | Plant-Doctor | 74.93±0.22 | 76.20±0.34 | 73.80±0.52 | 89.51±0.34 | 90.43±0.32 | 15±1 |
| | LeafTrackNet | **87.33**±0.61 | **86.50**±0.74 | **88.24**±0.51 | **93.69**±0.88 | **96.43**±0.68 | 8±5 |
| CanolaTrack | BoT-SORT | 82.39±0.20 | 83.71±0.46 | 81.19±0.27 | 93.41±0.65 | 93.02±0.27 | 6±1 |
| | ByteTrack | 76.69±0.37 | 77.35±0.23 | 76.20±0.59 | 92.03±0.80 | 92.70±0.46 | 7±1 |
| | MOTRv2 | **88.13**±0.22 | **87.32**±0.18 | **89.01**±0.28 | **94.61**±0.41 | **97.17**±0.25 | **1**±1 |
| | MOTIP | 67.70±4.52 | 84.08±0.63 | 54.75±6.94 | 64.42±3.06 | 63.79±5.15 | 278±52 |
| | LeTra | 76.50±0.42 | 77.78±0.36 | 75.44±0.58 | 86.01±0.36 | 86.10±0.79 | 22±2 |
| | Plant-Doctor | 75.00±2.15 | 76.27±0.15 | 73.90±4.06 | 90.03±0.26 | 90.91±3.21 | 16±1 |
| | LeafTrackNet | 87.56±0.70 | 86.69±0.59 | 88.51±0.88 | 94.07±0.62 | 96.69±0.54 | 4±1 |

MOTRv2 pretrained on CanolaTrack achieves the highest HOTA on KOMATSUNA.

it allows reliable association of individual leaves across days without relying on strong assumptions of smooth motion or spatial continuity.

To enable a direct visual comparison of the long-term tracking performance across methods, we computed binary accuracy heatmaps that highlight the tracking performance for a single plant (Fig. 6a) and for each individual leaf of the same plant (Fig. 6b). Fig. 6(a) demonstrates the impact of the embedding stability of our method on long-term tracking performance, showing high average accuracy across 31 days. The accuracy is defined as the proportion of leaves correctly detected and consistently tracked on a given day. While state-of-the-art methods exhibit significant performance degradation, especially after Day 9, our method maintains consistently high and stable accuracy throughout the plant's growth cycle. Fig. 6(b) provides a fine-grained view by visualizing the tracking accuracy of the individual leaves, where each cell reflects the success of identifying a specific leaf on a specific day. Yellow indicates a correct leaf association (IoU ≥ 0.75 and correct ID), purple denotes failure, and blank cells indicate the absence of leaves due to complete occlusion, senescence, or leaves that have not yet emerged. These visualizations demonstrate that LeafTrackNet provides more reliable long-term tracking in terms of both average daily accuracy and individual leaf trajectories.

### 4.3.2. KOMATSUNA

The KOMATSUNA dataset provides top-view image sequences of Komatsuna plants acquired from multiple views. We evaluate two training strategies for each tracker. In the first strategy, each method is fine-tuned directly on the KOMATSUNA training set from its default pretrained weights. In the second strategy, the same model is first fine-tuned on CanolaTrack and subsequently fine-tuned on the KOMATSUNA training set.

As shown in Table 4, LeafTrackNet achieves the highest HOTA of 87.33 under direct KOMATSUNA fine-tuning, outperforming MOTRv2 by 4.94 points. After sequential fine-tuning on CanolaTrack and KOMATSUNA, LeafTrackNet improves to 87.56 HOTA, whereas MOTRv2 increases more from 82.39 to 88.13 HOTA and exceeds LeafTrackNet by 0.57 points. MOTRv2 also records the fewest IDSW under both settings. These results provide two complementary findings. LeafTrackNet achieves the strongest HOTA performance without prior fine-tuning on CanolaTrack, while CanolaTrack provides transferable supervision improving MOTRv2. Under direct KOMATSUNA fine-tuning, MOTIP obtains 60.39 HOTA, substantially below its 79.68 HOTA on CanolaTrack. This decrease may result from its original settings being less suited to the smaller KOMATSUNA dataset. In particular, the 16 training sequences may provide insufficient trajectory diversity for adapting the identity association model, leading to repeated identity reassignments, as reflected by the relatively high IDSW count. Moreover, with only four test sequences, the average HOTA may be sensitive to a small number of association failures.

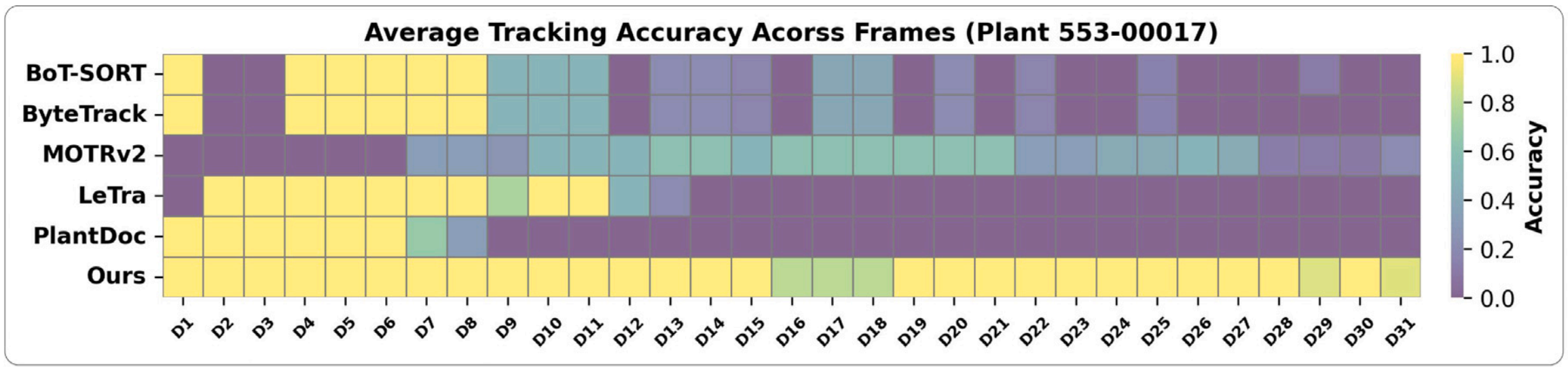

(a)

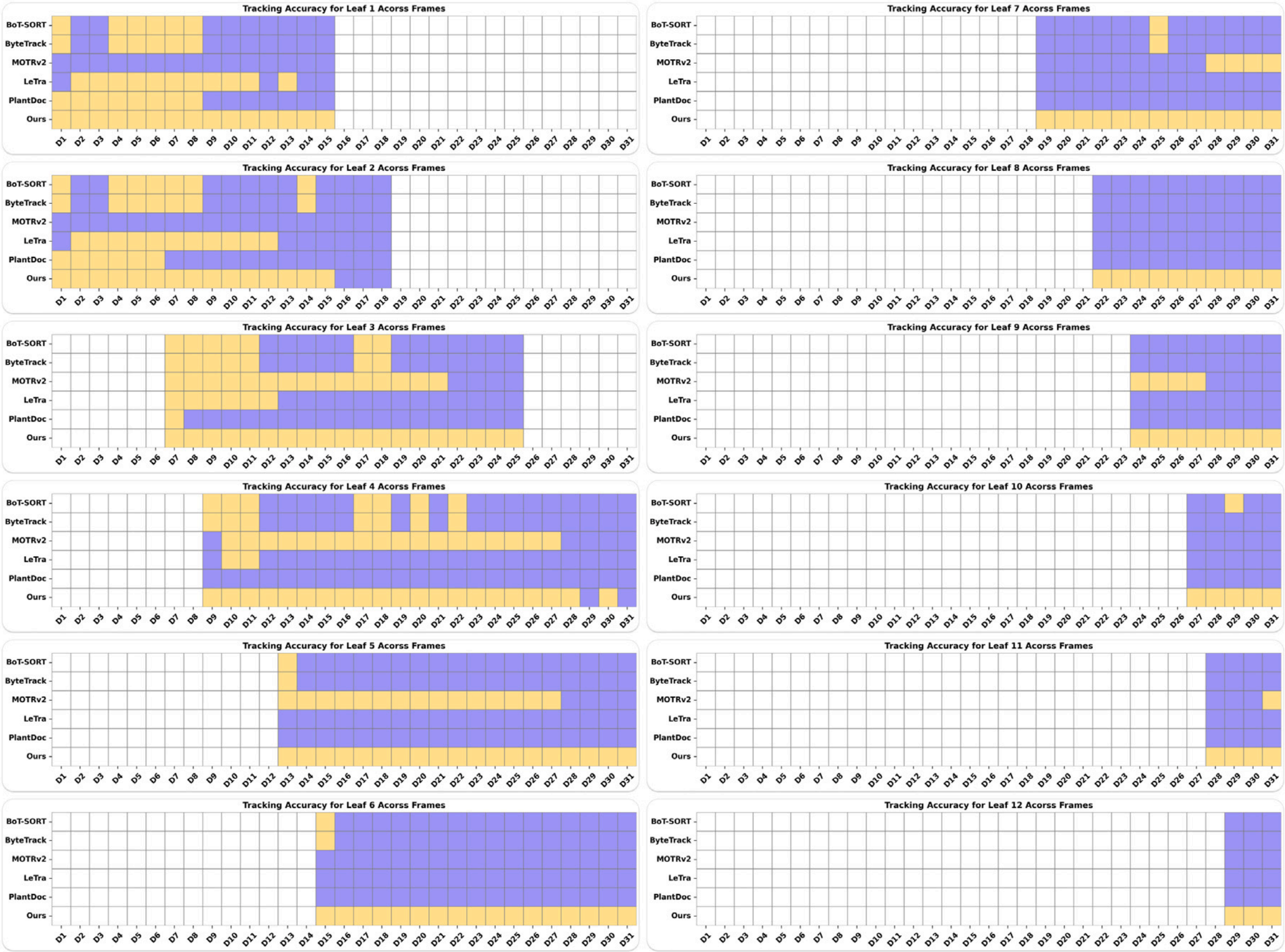

(b)

**Fig. 6.** Tracking accuracy visualization for Plant-158. (a) Average frame-level accuracy per method per day. (b) Per-leaf binary tracking matrix: yellow = correct, purple = failure, blank = leaf absent.

### 4.3.3. MSU-PID

MSU-PID is a synchronized multimodal dataset that includes Arabidopsis and bean plants captured with fluorescence, infrared, RGB, and depth imaging. We apply the same two training strategies used for KOMATSUNA and report the results in Table 5. Under direct MSU-PID fine-tuning from the default pretrained weights, LeafTrackNet achieves the highest HOTA of 74.20, outperforming MOTRv2 by 1.62 points. Its higher DetA compensates for the higher AssA of MOTRv2, resulting in better overall HOTA, although MOTRv2 records the fewest IDSW. Following sequential fine-tuning on CanolaTrack and MSU-PID, MOTRv2 improves from 72.58 to 74.82 HOTA and exceeds LeafTrackNet by 0.62 points. This reversal shows that MOTRv2 benefits more strongly from the additional CanolaTrack fine-tuning stage. MOTIP, by comparison, records high IDSW counts under both settings and large variation across random seeds, indicating unstable identity association on this multimodal dataset.

Taken together with the CanolaTrack and KOMATSUNA results, LeafTrackNet achieves the highest HOTA under direct fine-tuning on all three datasets. Notably, on CanolaTrack, the largest high-resolution crop leaf-tracking dataset to our knowledge, LeafTrackNet outperforms the strongest competing method by 8.35 HOTA points and MOTRv2 by 9.73 points. The sequential fine-tuning results provide a complementary finding that CanolaTrack offers transferable supervision that can substantially improve other tracking architectures, particularly MOTRv2.

**Table 5**
Tracking performance on the MSU-PID test set with direct MSU-PID fine-tuning (upper block) and sequential CanolaTrack-to-MSU-PID fine-tuning (lower block). Values are reported as mean ± standard deviation over three runs.

| Pretraining | Method | HOTA↑ | DetA↑ | AssA↑ | MOTA↑ | IDF1↑ | IDSW↓ |
|---|---|---|---|---|---|---|---|
| Default | BoT-SORT | 70.32±0.45 | 70.59±0.35 | 70.27±0.57 | **84.66**±0.54 | 86.35±0.49 | 70±2 |
| | ByteTrack | 68.29±0.43 | 66.85±0.32 | 70.09±0.57 | 83.10±0.14 | **88.60**±0.29 | 41±2 |
| | MOTRv2 | 72.58±1.03 | 66.01±1.41 | **80.18**±0.58 | 73.46±1.73 | 85.88±0.83 | **21**±1 |
| | MOTIP | 64.60±1.95 | 72.48±0.61 | 57.99±3.14 | 64.98±3.69 | 70.09±3.23 | 791±120 |
| | LeTra | 67.44±0.27 | 66.42±0.46 | 68.85±0.48 | 75.25±0.37 | 79.36±0.53 | 167±6 |
| | Plant-Doctor | 63.82±0.49 | 66.35±0.20 | 61.62±0.80 | 77.87±0.59 | 79.52±0.96 | 207±9 |
| | LeafTrackNet | **74.20**±0.47 | **74.09**±0.06 | 74.70±0.97 | 83.98±0.20 | 87.77±0.81 | 137±11 |
| CanolaTrack | BoT-SORT | 70.20±0.36 | 70.35±0.20 | 70.26±0.74 | **84.13**±0.30 | 86.40±0.85 | 68±6 |
| | ByteTrack | 67.95±0.64 | 66.54±0.37 | 69.69±1.05 | 82.27±0.99 | 87.99±1.21 | 47±16 |
| | MOTRv2 | **74.82**±0.74 | 70.21±0.57 | **80.07**±0.96 | 80.10±0.72 | **89.14**±0.70 | **22**±5 |
| | MOTIP | 66.92±7.16 | 72.85±0.48 | 62.28±12.53 | 68.85±8.46 | 74.01±10.89 | 664±321 |
| | LeTra | 68.18±0.29 | 67.20±0.05 | 69.55±0.59 | 75.88±0.14 | 79.79±0.27 | 170±4 |
| | Plant-Doctor | 64.23±0.31 | 66.25±0.17 | 62.51±0.59 | 77.69±0.25 | 80.52±0.46 | 204±1 |
| | LeafTrackNet | 74.20±0.65 | **74.04**±0.22 | 74.77±1.43 | 83.70±0.57 | 88.11±1.44 | 148±17 |

MOTRv2 pretrained on CanolaTrack achieves the highest HOTA on MSU-PID.

**Table 6**
Detector ablation on CanolaTrack. Metrics are reported as mean ± standard deviation over three runs. Best values are **bold**; second-best values are underlined. MACs are computed for 640 × 640 inputs.

| | Params(M) | MACs(G) | HOTA↑ | DetA↑ | AssA↑ | MOTA↑ | IDF1↑ |
|---|---|---|---|---|---|---|---|
| YOLO10x | 29.5 | 80.2 | **88.03**±0.24 | **92.25**±0.03 | **84.07**±0.49 | **93.64**±0.18 | **92.90**±0.35 |
| YOLO11x | 56.9 | 97.5 | 87.64±0.08 | 92.04±0.04 | 83.52±0.14 | 92.73±0.09 | 92.41±0.11 |
| YOLO12x | 59.1 | 99.5 | 87.07±0.07 | 91.58±0.05 | 82.86±0.08 | 92.03±0.06 | 91.82±0.05 |
| YOLO26x | 55.7 | 97.0 | 87.87±0.13 | 92.12±0.05 | 83.89±0.26 | 93.41±0.15 | 92.82±0.08 |

**Table 7**
Backbone ablation on CanolaTrack. Metrics are reported as mean ± standard deviation over three runs. Best values are **bold**; second best are underlined. MACs are computed for 224 × 224 inputs.

| | Params(M) | MACs(G) | HOTA↑ | DetA↑ | AssA↑ | MOTA↑ | IDF1↑ |
|---|---|---|---|---|---|---|---|
| MobileNetV3 | 2.97 | 0.23 | **88.03**±0.24 | 92.25±0.03 | **84.07**±0.49 | 93.64±0.18 | **92.90**±0.35 |
| ResNet18 | 11.18 | 1.82 | 87.67±0.62 | 92.22±0.01 | 83.41±1.18 | **93.65**±0.13 | 92.55±0.72 |
| ResNet34 | 21.28 | 3.68 | 87.31±1.29 | 92.28±0.06 | 82.69±2.40 | 93.53±0.80 | 91.98±1.21 |
| ResNet50 | 23.51 | 4.13 | 87.70±0.04 | 92.27±0.03 | 83.44±0.10 | 93.60±0.16 | 92.45±0.13 |
| ResNet101 | 42.50 | 7.86 | 87.10±0.47 | **92.29**±0.07 | 82.28±0.82 | 93.49±0.43 | 91.79±0.55 |
| ViT_B/16 | 85.80 | 17.61 | 86.79±0.75 | 92.24±0.03 | 81.75±1.43 | 92.97±0.42 | 91.61±0.88 |

### *4.4. Ablation study*

**Detectors.** Table 6 presents a detector ablation in which the default YOLOv10 detector is replaced with more recent YOLO variants. All detector–tracker configurations were trained and evaluated under identical experimental settings, including the same number of training epochs, optimizer and learning-rate schedule, batch size, data splits, and random seeds. Notably, replacing YOLOv10 with YOLOv11 (Khanam and Hussain, 2024), YOLOv12 (Tian et al., 2025), or YOLOv26 (Sapkota et al., 2026) does not improve tracking performance. The best overall results are achieved with YOLOv10 (e.g., 88.03 HOTA), while the alternative detectors exhibit slightly lower tracking accuracy. Since all evaluated detectors already provide strong detection performance on CanolaTrack, these results suggest a saturation effect within the tested detector range, rather than showing that detector capacity is generally irrelevant. YOLOv10 is therefore retained as the detector in our final pipeline because it achieves the best overall tracking performance among the evaluated variants.

**Backbone.** In Table 7, we evaluate the impact of different backbone architectures by replacing the default MobileNetV3 with different variants of ResNet (He et al., 2016) (ResNet18, ResNet34, ResNet50, ResNet101) and a Vision Transformer (Dosovitskiy et al., 2021) (ViT-B/16). Under identical training settings and detectors, MobileNetV3 achieves the strongest identity metrics (HOTA/AssA/IDF1) while using only approximately 3 million parameters. DetA remains nearly constant across backbones, as expected with a shared detector. Higher model capacity does not directly translate to better identity maintenance for structured, non-rigid leaf motion. Deeper ResNets and ViT-B/16 increase computation by 4–30× without improving tracking. ViT-B/16, despite its higher capacity, also underperforms, likely due to its reliance on large-scale data for effective generalization. These results suggest that compact backbones like MobileNetV3 are sufficient to efficiently learn discriminative, temporally stable embeddings.

**Triplet Sampling Strategy.** We evaluate the impact of three different triplet sampling strategies using MobileNetV3 as the backbone: (i) *cross-plant flexible sampling*, (ii) *intra-plant full-cycle sampling*, and (iii)

**Table 8**
Ablation of triplet sampling strategies and temporal window size ($\Delta T$) on CanolaTrack.

| Strategy | $\Delta T$ | HOTA↑ | DetA↑ | AssA↑ | MOTA↑ | IDF1↑ |
|---|---|---|---|---|---|---|
| (i) cross-plant flexible | – | **88.30**±0.24 | **92.25**±0.02 | **84.59**±0.46 | **94.13**±0.23 | **93.15**±0.14 |
| (ii) intra-plant full-cycle | – | 88.03±0.24 | 92.25±0.03 | 84.07±0.49 | 93.64±0.18 | 92.90±0.35 |
| (iii) intra-plant temporal windows | 1 | 59.29±0.67 | 92.23±0.06 | 38.18±0.88 | 83.59±0.56 | 59.79±1.00 |
| | 2 | 64.83±0.58 | 92.14±0.05 | 45.69±0.79 | 86.29±0.12 | 65.09±0.37 |
| | 5 | 72.91±8.25 | 92.22±0.03 | 58.19±13.42 | 88.09±2.86 | 74.76±9.92 |
| | 10 | 64.96±1.48 | 92.16±0.08 | 45.84±2.09 | 84.02±0.80 | 65.43±2.13 |
| | 20 | 59.10±4.13 | 92.14±0.07 | 38.07±5.22 | 80.17±2.91 | 58.02±5.75 |

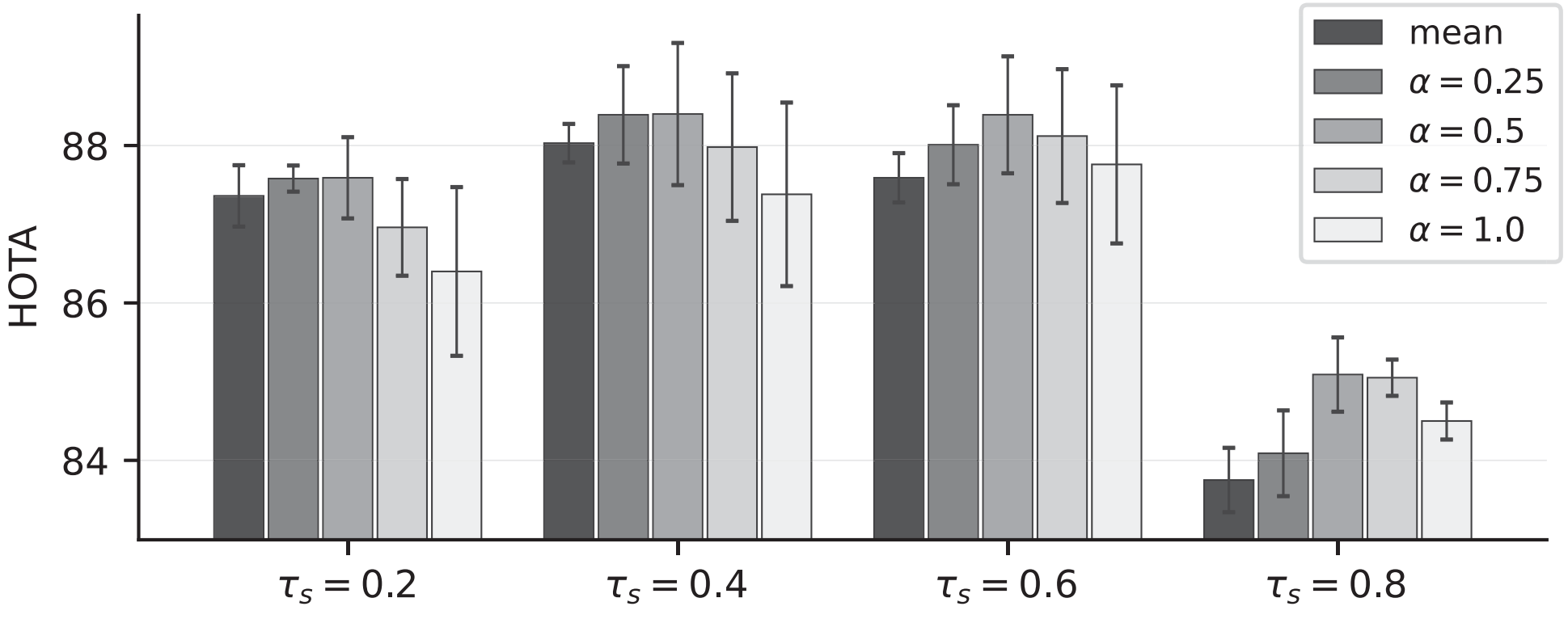

**Fig. 7.** Inference ablation of the similarity threshold $\tau_s$ and temporal smoothing coefficient $\alpha$. Error bars indicate ± one standard deviation over three training runs.

*intra-plant temporal-window sampling*. In *cross-plant flexible sampling*, the anchor and positive samples are selected from the same plant, and the negative sample is randomly drawn from the entire training set without a plant constraint. In the *intra-plant full-cycle sampling* strategy, all three samples are drawn randomly from the same plant across the entire 31-day sequence. In *intra-plant temporal-window sampling*, all samples are drawn from the same plant, but the negative sample is restricted to a $\Delta T$-day neighborhood around the anchor.

As shown in Table 8, cross-plant flexible sampling achieves the highest HOTA score of 88.30, slightly surpassing our currently adopted intra-plant full-cycle sampling, suggesting that allowing negative samples from a broader distribution strengthens contrastive supervision. The intra-plant full-cycle strategy performs similarly (HOTA 88.03), indicating that sampling throughout the growth cycle of a single plant already offers consistent and informative embedding patterns that generalize well to long-term appearance variation. In contrast, intra-plant temporal-window sampling exhibits substantial degradation in performance, with the largest drop at small $\Delta T$ and again at large $\Delta T$. Small windows yield easy negatives that provide little discriminative pressure within a rosette, while very large windows bias training toward trivially separable pairs. The performance of temporal-window sampling remains consistently lower than the other sampling strategies. Although cross-plant sampling achieves a slightly higher HOTA, we adopt intra-plant sampling in our final model because inference is performed within each plant sequence, where each leaf is associated with candidates from the same plant rather than from different plants.

**Inference hyperparameters.** We ablate the similarity threshold $\tau_s$ and the EMA coefficient $\alpha$ (Fig. 7). HOTA forms a plateau at $\tau_s \in [0.4, 0.6]$ and drops at $\tau_s = 0.8$ (over-pruning) and $\tau_s = 0.2$ (noisy associations). Within each $\tau_s$, performance improves with moderate smoothing and declines for $\alpha = 0.25$ (history-dominated, slow to adapt) and $\alpha = 1.0$ (one-frame memory that overwrites history). The history *mean* baseline (uniform average of past embeddings) is consistently below EMA with $\alpha = 0.5$ for all $\tau_s$, indicating that equal weighting underemphasizes recent morphology. Error bars are small (standard deviation ≤ 1 HOTA) across seeds.

**Table 9**
Large-displacement pair evaluation on the CanolaTrack test set. The subset contains adjacent-frame pairs with normalized motion score $D_t > 2.00$, meaning that the median ground-truth leaf-centroid displacement is more than twice the test-set median. Pair-IDSW Rate is the number of identity switches normalized by the number of ground-truth identity transitions. Values are reported as mean±standard deviation over three runs.

| Method | Pair-HOTA ↑ | Pair-DetA ↑ | Pair-AssA ↑ | Pair-IDSW Rate ↓ |
|---|---|---|---|---|
| BoT-SORT | 64.20±0.09 | 57.69±0.14 | 74.59±0.02 | 14.05±0.24 |
| ByteTrack | 59.61±0.28 | 52.78±0.23 | 71.96±0.25 | 14.10±0.16 |
| MOTRv2 | 76.47±2.06 | 67.09±3.29 | 87.57±1.10 | **6.46**±1.20 |
| MOTIP | 88.29±0.17 | 88.91±0.08 | 88.07±0.30 | 17.58±0.57 |
| LeTra | 78.09±0.21 | 81.48±0.20 | 75.67±0.24 | 39.32±0.52 |
| Plant-Doctor | 70.33±0.22 | 72.55±0.31 | 70.95±0.27 | 41.72±0.24 |
| LeafTrackNet | **90.48**±0.22 | **92.44**±0.02 | **88.86**±0.42 | 15.44±0.81 |

**Displacement-stress pair evaluation.** To quantify tracking behavior under abrupt frame-to-frame layout changes, including pot rotation, we constructed a large-displacement subset from the CanolaTrack test set. For each adjacent frame pair $(t, t+1)$, we computed $d_t$ as the median Euclidean centroid displacement of ground-truth leaf identities visible in both frames. This value was then normalized by the median displacement over all adjacent test-frame pairs, giving the normalized motion score $D_t$. Table 9 reports results for the challenging subset defined by $D_t > 2.00$, meaning that the median leaf displacement in the frame pair is more than twice the test-set median. This subset contains 94 adjacent-frame pairs and 624 ground-truth identity transitions. Pair metrics are computed on adjacent pairs extracted from the full-sequence predictions. Therefore, they measure local transition robustness across one difficult frame-to-frame step, rather than long-term identity preservation over complete plant growth sequences as in Table 3. Table 9 shows that LeafTrackNet achieves the highest Pair-HOTA, Pair-DetA, and Pair-AssA on this large-displacement subset, indicating strong local detection and association under abrupt layout changes. MOTIP performs competitively in this pairwise setting, with a smaller gap to LeafTrackNet than in the full-sequence evaluation. Specifically, the Pair-HOTA gap between LeafTrackNet and MOTIP is 2.19 points on the large-displacement subset, while the full-sequence HOTA gap is 8.35 points in Table 3. This pattern is expected because pairwise evaluation focuses on immediate adjacent-frame correspondence and does not capture accumulated identity drift, fragmentation, or recovery after longer occlusions. However, LeafTrackNet shows a clear advantage in both the pairwise setting and complete growth sequences by maintaining identity consistency. The full threshold sweep in Table A.1 further confirms that LeafTrackNet remains robust across increasing displacement thresholds.

## 5. Limitations

LeafTrackNet's tracking accuracy remains sensitive to the reliability of the upstream detector. False negatives can fragment trajectories, while duplicate detections and slight bounding-box shifts may propagate to association errors in dense rosettes. Although the proposed association module improves identity consistency, failures can still occur under severe inter-leaf occlusion and during reappearance, particularly when only a small and visually ambiguous leaf portion is visible. Tracking performance may also be sensitive to the age threshold $\tau_a$, which determines how long unmatched tracks remain in the memory bank. In this study, $\tau_a$ was fixed at 5 across all datasets without dataset-specific tuning, and the 29.44-point HOTA difference between the embedding-only variants using $\tau_a = 5$ and $\tau_a = 0$ indicates that broader evaluation of this parameter is needed in the future.

In addition, our supervised association training relies on per-leaf identity annotations, which are expensive to obtain and can limit rapid scaling to new crops in real phenotyping workflows. The current pipeline represents leaves using top-down 2D bounding boxes. Incorporating instance masks would isolate leaf pixels more precisely, reduce background in the extracted crops, and yield cleaner leaf appearance embeddings. Moreover, while LeafTrackNet is architecturally crop-agnostic in principle, it currently benefits from training or fine-tuning on the target domain. The transfer experiments suggest that transformer-based trackers benefit markedly from additional large-scale pretraining. For example, MOTRv2 improves substantially after pretraining on CanolaTrack and marginally outperforms LeafTrackNet by 0.57 HOTA on KOMATSUNA and 0.62 HOTA on MSU-PID. This observation highlights a practical constraint in agricultural phenotyping, where collecting and annotating leaf-level tracking data at scale is costly and often infeasible. In contrast, LeafTrackNet maintains strong performance under limited target-domain data, indicating better data efficiency in realistic low-data settings. The current evaluation is limited to reporting mean ± standard deviation over three independent runs without statistical significance testing and should therefore be interpreted descriptively. A robust statistical analysis would require additional independent runs and more extensive computational resources. Future work should validate these trends on larger and more diverse benchmarks and explore foundation-model features and self-supervised adaptation to improve cross-crop, cross-view, and cross-modality generalization, with the long-term goal of robust zero-shot leaf tracking under distribution shifts.

## 6. Conclusion

In this paper, we propose LeafTrackNet, a robust deep learning framework for leaf tracking from top-down RGB sequences of canola plants. By combining a high-accuracy leaf detector with a memory-based embedding association strategy, it effectively addresses core biological and environmental challenges such as leaf emergence, occlusion, deformation, and rotation. To support this development, we present CanolaTrack, a large-scale, high-resolution dataset comprising images of 184 plants collected over 31 days that offers a new benchmark for complex, structured, and long-term leaf tracking. Our proposed method enables accurate, scalable, and temporally consistent leaf identity tracking, which is an essential step toward fine-grained, automated plant phenotyping. While the current results demonstrate robust tracking in controlled laboratory phenotyping settings, validation under field conditions remains essential because of domain shifts caused by dense canopy structure, severe inter-leaf occlusion, background clutter, and illumination variation. This work provides a methodological foundation for future research in cross-species transferability, field-oriented validation, and real-time agricultural decision-making.

## CRediT authorship contribution statement

**Shanghua Liu:** Writing – review & editing, Writing – original draft, Visualization, Validation, Software, Methodology, Investigation, Formal analysis, Data curation, Conceptualization. **Majharulislam Babor:** Writing – review & editing, Writing – original draft, Supervision. **Christoph Verduyn:** Writing – review & editing, Supervision, Resources, Data curation. **Breght Vandenberghe:** Writing – review & editing, Data curation, Conceptualization. **Bruno Betoni Parodi:** Writing – review & editing, Project administration, Conceptualization. **Cornelia Weltzien:** Writing – review & editing, Conceptualization. **Marina M.-C. Höhne:** Writing – review & editing, Writing – original draft, Visualization, Validation, Supervision, Methodology, Funding acquisition, Formal analysis, Conceptualization.

## Declaration of competing interest

The authors declare that they have no known competing financial interests or personal relationships that could have appeared to influence the work reported in this paper.

## Acknowledgment

This work was funded by the Federal Ministry of Research, Technology and Space through the projects DCropS4OneHealth (ref. 16LW0528K) and REFRAME (ref. 01IS24073B).

## Appendix. Displacement threshold evaluation

See Table A.1.

## Data availability

Our dataset, code, and trained model weights are all publicly available at https://github.com/shl-shawn/LeafTrackNet.

**Table A.1**
Threshold sweep for the large-displacement pair evaluation on the CanolaTrack test set. Adjacent-frame subsets are defined by increasing thresholds on the normalized motion score $D_t$, computed from the median ground-truth leaf centroid displacement of each frame pair. Values are reported as mean±standard deviation over three runs. The shaded $D_t > 2.00$ column corresponds to Table 9.

| Method | All adjacent pairs | $D_t > 1.00$ | $D_t > 1.50$ | $D_t > 2.00$ | $D_t > 3.00$ | $D_t > 4.00$ |
|---|---|---|---|---|---|---|
| **Subset size** | | | | | | |
| #Pairs | 1102 | 551 | 213 | 94 | 41 | 27 |
| #GT Trans. | 5791 | 3520 | 1376 | 624 | 280 | 206 |
| **(a) Pair-HOTA ↑** | | | | | | |
| BoT-SORT | 75.35±0.02 | 74.56±0.03 | 70.83±0.05 | 64.20±0.09 | 50.34±0.22 | 47.59±0.24 |
| ByteTrack | 69.51±0.11 | 69.19±0.08 | 65.95±0.24 | 59.61±0.28 | 47.57±0.15 | 44.76±0.33 |
| MOTRv2 | 86.61±1.59 | 84.83±2.04 | 82.34±2.29 | 76.47±2.06 | 67.98±2.77 | 63.73±2.64 |
| MOTIP | 91.09±0.10 | 91.45±0.12 | 90.28±0.17 | 88.29±0.17 | 84.58±0.63 | **83.56±0.72** |
| LeTra | 87.09±0.08 | 86.81±0.16 | 83.66±0.21 | 78.09±0.21 | 66.82±0.01 | 61.80±0.18 |
| Plant-Doctor | 76.99±0.07 | 76.62±0.03 | 74.24±0.09 | 70.33±0.22 | 63.12±0.20 | 60.24±0.09 |
| LeafTrackNet | **94.17±0.08** | **94.39±0.11** | **93.20±0.08** | **90.48±0.22** | **85.78±0.59** | 83.29±0.63 |
| **(b) Pair-DetA ↑** | | | | | | |
| BoT-SORT | 66.79±0.05 | 66.12±0.07 | 62.48±0.03 | 57.69±0.14 | 49.49±0.28 | 50.69±0.21 |
| ByteTrack | 59.95±0.12 | 59.87±0.09 | 56.96±0.30 | 52.78±0.23 | 46.21±0.20 | 46.98±0.12 |
| MOTRv2 | 78.05±2.71 | 75.24±3.43 | 72.90±3.88 | 67.09±3.29 | 61.65±4.49 | 60.13±5.07 |
| MOTIP | 89.12±0.16 | 90.02±0.19 | 89.44±0.12 | 88.91±0.08 | 87.83±0.23 | 88.89±0.30 |
| LeTra | 83.01±0.14 | 83.71±0.27 | 82.48±0.34 | 81.48±0.20 | 79.57±0.04 | 81.10±0.15 |
| Plant-Doctor | 72.58±0.15 | 72.73±0.05 | 72.31±0.19 | 72.55±0.31 | 73.24±0.29 | 74.73±0.18 |
| LeafTrackNet | **92.58±0.01** | **93.43±0.00** | **93.34±0.01** | **92.44±0.02** | **91.88±0.00** | **92.68±0.00** |
| **(c) Pair-AssA ↑** | | | | | | |
| BoT-SORT | 88.09±0.03 | 87.26±0.04 | 83.35±0.12 | 74.59±0.02 | 53.76±0.27 | 47.09±0.33 |
| ByteTrack | 86.12±0.06 | 85.46±0.10 | 81.17±0.17 | 71.96±0.25 | 52.70±0.43 | 46.14±0.60 |
| MOTRv2 | **96.46±0.26** | **95.94±0.37** | **93.34±0.61** | 87.57±1.10 | 75.41±1.90 | 67.98±1.76 |
| MOTIP | 93.60±0.33 | 93.29±0.26 | 91.50±0.41 | 88.07±0.30 | **81.90±1.02** | **78.93±1.10** |
| LeTra | 92.33±0.06 | 90.89±0.06 | 85.69±0.10 | 75.67±0.24 | 56.79±0.04 | 47.63±0.22 |
| Plant-Doctor | 85.09±0.08 | 84.04±0.12 | 79.30±0.13 | 70.95±0.27 | 56.86±0.17 | 50.94±0.23 |
| LeafTrackNet | 96.12±0.16 | 95.60±0.21 | 93.31±0.17 | **88.86±0.42** | 80.40±1.10 | 75.11±1.13 |
| **(d) Pair-IDSW Rate ↓** | | | | | | |
| BoT-SORT | 1.91±0.05 | 2.85±0.11 | 6.69±0.13 | 14.05±0.24 | 28.93±0.36 | 37.38±0.84 |
| ByteTrack | 1.88±0.03 | 2.83±0.04 | 6.66±0.15 | 14.10±0.16 | 28.57±0.36 | 37.22±0.74 |
| MOTRv2 | **0.84±0.10** | **1.30±0.19** | **3.20±0.55** | **6.46±1.20** | **13.93±2.70** | **18.61±4.19** |
| MOTIP | 7.70±0.73 | 8.67±0.65 | 11.96±0.99 | 17.58±0.57 | 27.02±1.25 | 32.52±0.97 |
| LeTra | 8.48±0.11 | 11.52±0.07 | 21.41±0.29 | 39.32±0.52 | 72.02±0.21 | 86.89±0.49 |
| Plant-Doctor | 12.38±0.05 | 14.84±0.03 | 24.39±0.04 | 41.72±0.24 | 71.43±0.62 | 83.50±0.00 |
| LeafTrackNet | 3.95±0.26 | 5.21±0.34 | 8.70±0.23 | 15.44±0.81 | 28.93±1.99 | 37.54±2.02 |

Pair-IDSW Rate is the number of identity switches normalized by the number of GT identity transitions.